\documentclass[conference, final]{IEEEtran}
\IEEEoverridecommandlockouts
\usepackage{cite}
\usepackage{amsmath,amssymb,amsfonts}
\usepackage{algorithmic}
\usepackage{graphicx}
\usepackage{textcomp}

\usepackage[inline]{enumitem}
\usepackage{algorithmic}
\usepackage{graphicx}
\usepackage{textcomp}
\usepackage{xcolor}

\usepackage{witharrows}

\usepackage[flushleft]{threeparttable}
\usepackage{tablefootnote}
\usepackage{array}

\newcolumntype{P}[1]{>{\centering\arraybackslash}m{#1}}
\usepackage{multicol}
\usepackage{multirow}
\usepackage{tabulary}
\usepackage{colortbl}
\definecolor{mygray}{gray}{0.90}
\usepackage{makecell}
\usepackage{booktabs}
\usepackage{subcaption}
\usepackage{stmaryrd}
\usepackage{float}
\usepackage{pifont}

\usepackage{arydshln}
\colorlet{mygray}{gray!15!white}

\usepackage[switch,columnwise]{lineno}

\usepackage{url,hyperref,microtype}
\hypersetup{
    colorlinks=true,
    citecolor=blue,
    linkcolor=blue,
    filecolor=magenta,      
    urlcolor=black,
}

\def\BibTeX{{\rm B\kern-.05em{\sc i\kern-.025em b}\kern-.08em
    T\kern-.1667em\lower.7ex\hbox{E}\kern-.125emX}}
\begin{document}

\title{MUPA\textsuperscript{2}S: Multimodal Unified Perception with Asymmetric Attention for Stress Assessment\\
}

\author{

\IEEEauthorblockN{Stefanos Gkikas}
\IEEEauthorblockA{\textit{Honda Research Institute Japan} \\
Wako City, Japan \\
stefanos.gkikas@jp.honda-ri.com}

\and

\IEEEauthorblockN{Thomas Kassiotis}
\IEEEauthorblockA{\textit{Department of Electronic Engineering} \\
\textit{Hellenic Mediterranean University}\\
Chania, Greece \\
ddk305@edu.hmu.gr}

\and

\IEEEauthorblockN{Yu Fang}
\IEEEauthorblockA{\textit{Honda Research Institute Japan} \\
Wako City, Japan  \\
yu.fang@jp.honda-ri.com}

\and

\IEEEauthorblockN{Christian Arzate Cruz}
\IEEEauthorblockA{\textit{Honda Research Institute Japan} \\
 Wako City, Japan  \\
 christian.arzate@jp.honda-ri.com}

\and

\IEEEauthorblockN{Miao Cheng}
\IEEEauthorblockA{\textit{Research Institute of Electrical Communication} \\
\textit{Tohoku University} \\
Sendai, Japan \\
cheng@tohoku.ac.jp}

\and

\IEEEauthorblockN{Zongyi Jiang}
\IEEEauthorblockA{
\textit{University of Canberra} \\
Canberra, Australia \\
zongyi.jiang@canberra.edu.au
}

\and

\IEEEauthorblockN{Zacharias Kamarianakis}
\IEEEauthorblockA{\textit{Department of Electronic Engineering} \\\
\textit{Hellenic Mediterranean University}\\
Chania, Greece \\
zkamar@hmu.gr}

\and

\IEEEauthorblockN{Houshyar Asadi}
\IEEEauthorblockA{\textit{College of Science \& Engineering}\\
\textit{Flinders University}\\
Adelaide, Australia \\
houshyar.asadi@flinders.edu.au}

\and

\IEEEauthorblockN{Randy Gomez}
\IEEEauthorblockA{\textit{Honda Research Institute Japan} \\
Wako City, Japan \\
r.gomez@jp.honda-ri.com}

\and

\IEEEauthorblockN{Raul Fernandez Rojas}
\IEEEauthorblockA{\textit{BioSIS (Biosensing \& Intelligent Systems) Lab} \\
\textit{Centre for Intelligent Computing and Systems} \\
\textit{University of Canberra} \\
Canberra, Australia \\
raul.fernandezrojas@canberra.edu.au}

\and

\IEEEauthorblockN{Giorgos Giannakakis}
\IEEEauthorblockA{\textit{Department of Electronic Engineering} \\
\textit{Hellenic Mediterranean University}\\
Chania, Greece \\
ggian@hmu.gr}

}

\maketitle

\begin{abstract}
Stress is a leading contributor to long-term physical and mental health decline. The automatic assessment of stress is a promising approach for continuous monitoring and mitigating human biases. 
However, the development of automated systems remains challenging due to the heterogeneity of physiological and behavioral signals.
This paper presents MUPA\textsuperscript{2}S, a multimodal framework for automatic stress assessment that combines 
facial video and ECG signals within a unified perception architecture. Both modalities are mapped into a shared 
token representation without modality-specific components, and processed through asymmetric attention comprising segment-local cross-attention and global self-attention across segment states.
Experiments on a newly collected $58$-subject stress dataset, evaluated under a stratified subject-level protocol, show that facial video achieves $69.44\%$ test accuracy, while ECG-only performance remains close to chance under the proposed representation. Unified Video--ECG fusion achieves a test accuracy of $69.58\%$ at approximately half the number of parameters of feature-level fusion, with negligible additional computational cost. Feature-level fusion does not exceed the video-only baseline, while unified input-level integration provides a parameter-efficient mechanism for jointly processing the two modalities.
\end{abstract}

\begin{IEEEkeywords}
Stress recognition, mental health, affective computing, electrocardiography, transformer 
\end{IEEEkeywords}


\section{Introduction}

Stress is a physiological and psychological response to perceived demands, involving coordinated autonomic nervous system activity and neuroendocrine stress responses \cite{goldstein_2023}. It manifests across a spectrum from acute, time-limited episodes to chronic, sustained states, each with distinct physiological signatures and implications for health. Self-report instruments such as the Perceived Stress Scale remain widely used for assessing perceived stress \cite{cohen_kamarck_1983}, yet global retrospective self-reports are limited by recall bias and are not well suited to capturing how stress levels change over time and across contexts \cite{shiffman_stone_2008}. 


Stress is a growing global health concern. Analysis of nationally representative surveys across $146$ countries found that the likelihood of stress reporting doubled over $18$ years, with increasing disparities across gender, age, and income groups \cite{canaletti_lun_2026}. Furthermore, psychosocial work exposures contribute substantially to cardiovascular disease and depression across $28$ European countries \cite{sultantaib_villeneuve_2022}. 
In settings requiring continuous or unobtrusive monitoring, conventional stress assessment remains limited. Self-reports may be impractical, delayed, or biased, whereas contact-based wearable systems face compliance, motion-artifact, and scalability challenges \cite{hosseini_gottumukkala_2026}. Moreover, naturalistic deployments reveal persistent issues with data quality and signal integrity \cite{neigel_vargo_2025}. These limitations motivate automated, non-intrusive systems capable of passively and objectively recognizing stress from readily available signals \cite{giannakakis_grigoriadis_2022}.

Stress recognition has been studied across physiological and behavioral modalities. ECG and EDA reflect autonomic arousal, although inter-individual variability limits generalization \cite{gonzalez_carabarin_2021}. Facial behavior offers a contact-free alternative through eye, mouth, and head activity \cite{giannakakis_pediaditis_2017} or Facial Action Units \cite{giannakakis_koujan_2020, kassiotis_stressgat_acii_2026}. Deep learning \cite{kyrou_kompatsiaris_2025} and multimodal fusion \cite{zhang_yin_2022} have further improved recognition. Facial video and biosignals were recently combined for driver stress estimation \cite{valergaki_nicodemou_2026}, using 3D facial descriptors with cross-modal attention. In contrast, the present study maps facial video frames and ECG into a shared representation processed by a single asymmetric-attention backbone without modality-specific encoders, and is evaluated across multiple stress-induction conditions rather than a driving protocol.

\section{Related Work}
\label{related_work}

Early work on video-based stress recognition relied on handcrafted facial cues, including eye activity, mouth movements, head-motion parameters, and camera-derived heart rate information, demonstrating that these cues can distinguish stressed and anxious states from neutral or relaxed ones \cite{giannakakis_pediaditis_2017}. The adoption of deep learning shifted video-based stress recognition toward learned face- and action-level representations, outperforming handcrafted feature-engineering approaches on a video-based stress dataset \cite{zhang_feng_2020}. Spatiotemporal deep architectures have been proposed to improve facial image-based stress recognition by modeling facial information across regions and over time \cite{jeon_bae_2021}. Facial Action Units provide an interpretable intermediate representation for stress recognition; AU-based models have been applied to distinguish stressed from neutral states in a contact-free manner \cite{giannakakis_koujan_2020}. More recently, the integration of explainable AI techniques into AU-based stress recognition has enabled identification of the specific AUs that drive predictions, improving both model transparency and physiological interpretability \cite{giannakakis_roussos_2025}.

Physiological signals provide complementary indices of stress-related autonomic activity and are less directly dependent on overt facial behavior \cite{giannakakis_grigoriadis_2022}. Heart rate variability derived from ECG has been widely studied as a marker of autonomic stress regulation \cite{giannakakis_grigoriadis_2022}; pairwise transformations and feature-selection strategies applied to HRV features have been used to improve stress classification performance \cite{giannakakis_marias_2019}. ECG has also been studied in biometric recognition under different psychological stress states, showing that stress-related changes in cardiac dynamics affect subject-specific ECG representations and motivate identity-aware modeling \cite{zhou_wang_2021}. Multi-signal approaches combining EEG and ECG highlight the challenge of inter-individual variability in physiological stress responses and motivate personalized machine learning strategies for more generalizable models \cite{gonzalez_carabarin_2021}. Despite high accuracy in controlled settings, wearable affect-monitoring systems face challenges in unconstrained environments, where motion artifacts and signal quality degradation limit deployment in practice \cite{alkurdi_he_2025}.

Given the limitations of each modality considered in isolation, multimodal approaches have attracted growing interest as fusion frameworks can exploit complementary physiological and behavioral information for stress recognition. A hybrid deep learning and handcrafted-feature fusion system that combines ECG and EDA signals has been shown to outperform unimodal baselines in a leave-one-out cross-validation on the WESAD dataset \cite{zhao_niu_2023}. Deep learning architectures fusing ECG, facial expressions, and voice have demonstrated that matrix eigenvector-based cross-modal fusion captures acute stress dynamics more effectively than individual streams \cite{zhang_yin_2022}. In a simulated office environment, the combination of cardiac, mouse, and keyboard features with interpretable machine learning has enabled multi-level stress detection, showing that behavioral interaction signals carry stress-related information beyond physiological channels alone \cite{naegelin_weibel_2023}. The fusion of facial video and cardiovascular signals has recently been explored for stress estimation in diverse scenarios, including driving, where cross-modal attention mechanisms align temporal facial and physiological dynamics during distracted-driving stressors \cite{valergaki_nicodemou_2026}. Comprehensive surveys of deep learning for stress detection identify multimodal fusion as a prominent direction for improving robustness and performance across heterogeneous datasets and protocols \cite{kyrou_kompatsiaris_2025}.

Regarding the modeling of stress, early approaches relied predominantly on classical machine learning, where statistical, temporal, and frequency-domain features were extracted from the signals and fed to classifiers such as Decision Trees, Random Forests, AdaBoost, and Linear Discriminant Analysis \cite{schmidt_reiss_2018}, while the recognized physical activity of the user has been incorporated as context to distinguish psychological stress from similar physiological arousal in daily-life settings \cite{gjoreski_lustrek_2017}. With the adoption of deep learning, end-to-end architectures based on convolutional and recurrent layers have been employed to learn representations directly from raw physiological signals, achieving results comparable to or better than feature-based pipelines while removing the need for manual feature engineering \cite{dziezyc_gjoreski_2020}. Furthermore, self-supervised learning has been introduced to reduce the dependence on labeled data, with networks pre-trained to recognize transformations applied to unlabeled ECG signals and subsequently transferred to the classification of stress and other affective states \cite{sarkar_etemad_2022}. More recently, transformer-based architectures have been utilized, either combining convolutional layers with a transformer encoder to detect stress directly from ECG \cite{behinaein_bhatti_2021}, or employing a shared transformer encoder pre-trained in a self-supervised manner to capture correlations within and across multiple wearable signals \cite{wu_daoudi_2024}.


\section{Methodology}
\label{sec:methodology}

\subsection{Unified Perception}

The framework maps heterogeneous inputs into a shared token representation without modality-specific architectures, enabling a single model to process both behavioral and physiological signals. For a video sequence of $L$ frames, each frame is a $224\times224$ RGB image. Rather than processing frames sequentially, the temporal dimension is folded into the channel dimension---a step referred to as \textit{axis folding}---creating a tensor of shape $H\times W\times 3L$. This preserves the spatial structure of each frame while packing the temporal structure into a single 2D representation. For ECG, the 1D waveform is first mean-centered and standardized, min--max normalized to $[0,1]$, represented as a $1\times L$ grid, and bilinearly interpolated to $224\times224$, yielding a single-channel 2D representation. The two representations are concatenated along the channel dimension:
\begin{equation}
\mathbf{X} \in \mathbb{R}^{B \times H \times W \times (3L+1)},
\end{equation}
where $B$ denotes the batch size and $H=W=224$.
Geometric information is incorporated by encoding each spatial position $\mathbf{p}\in[-1,1]^2$ using Fourier features. The model uses $K=6$ frequency bands and a maximum frequency $f_{\max}=10$. Since the input has $D=2$ spatial axes, the Fourier encoding adds $D(2K+1)=26$ positional features. The encoding is:
\begin{equation}
\begin{split}
\gamma(\mathbf{p}) = \bigl[&\sin(\pi s_1 \mathbf{p}),\ \cos(\pi s_1 \mathbf{p}),\ \ldots,\\
&\sin(\pi s_K \mathbf{p}),\ \cos(\pi s_K \mathbf{p}),\ \mathbf{p}\bigr],
\end{split}
\end{equation}
where $\{s_k\}_{k=1}^{K}$ spans $[1, f_{\max}/2]$. The spatial axes are flattened into a sequence of $N = H\times W = 50176$ tokens, with data channels and positional features concatenated per token to form the token matrix:
\begin{equation}
\mathbf{T} \in \mathbb{R}^{B \times N \times C'},\quad C' = (3L+1) + D(2K+1),
\end{equation}
where $D=2$ denotes the number of spatial axes. Since $D=2$ and $K=6$, the token dimension becomes:
\begin{equation}
C' = (3L+1) + 26 = 3L+27.
\end{equation}
The token sequence is partitioned into $S=4$ contiguous segments of length $n_s = N/S = 12544$. 
These settings are fixed across all experiments. Fourier features preserve spatial-position information after flattening the $224\times224$ representation, while $S=4$ divides the $50176$ tokens into four equal groups of $12544$ tokens for local cross-attention before global interaction among the segment states. The ECG interpolation maps the 1D waveform to the same $224\times224$ domain as the folded video, enabling direct channel-level integration within a single shared architecture.
The tokens of segment $s$ are denoted $\tilde{\mathbf{T}}_s \in \mathbb{R}^{B\times n_s\times C'}$.
Similar approaches of unified perception fusing heterogeneous modalities via interpolation can be found \cite{gkikas_pain_icmi_2026, gkikas_cruz_acii_cwl_2026}, showing their efficacy and efficiency for human state recognition.

\subsection{Asymmetric Attention}
\label{sec:asymmetric}
The model processes the segmented token sequence through four layers, each comprising a cross-attention block followed by $R_\ell$ self-attention blocks. A single latent state is associated with each segment, instantiated at runtime by replicating a shared initialization vector derived from a set of $M_0=32$ learnable global parameters $\{\boldsymbol{\ell}_m\}_{m=1}^{M_0}$:
\begin{equation}
\boldsymbol{\ell}_{\mathrm{init}} = \frac{1}{M_0}\sum_{m=1}^{M_0}\boldsymbol{\ell}_m \in \mathbb{R}^{d_0},
\end{equation}
where $d_0=128$. These segment states are not independently learnable; segment-specific representations emerge through the attention updates \cite{gkikas2026multiscale, gkikas2026mupa2e}.

\textbf{Cross-attention.} At each layer $\ell$, each segment state aggregates information exclusively from its corresponding token subset through cross-attention:
\begin{equation}
\mathbf{e}_s^{(\ell)} = \mathbf{e}_s^{(\ell-1)} + \mathrm{Attn}\bigl(\mathbf{e}_s^{(\ell-1)},\ \tilde{\mathbf{T}}_s\bigr),
\end{equation}
where $\mathbf{e}_s^{(\ell-1)}\in\mathbb{R}^{B\times 1\times d_\ell}$ provides the queries and $\tilde{\mathbf{T}}_s\in\mathbb{R}^{B\times n_s\times C'}$ provides the keys and values. This operation is asymmetric: the query side consists of a single vector of dimension $d_\ell$, while the key-value side spans $n_s \gg 1$ token vectors of dimension $C'$. The resulting attention matrix is $1\times n_s$, not square, and the query and key-value spaces differ in both size and dimensionality. All $S$ segments are processed in parallel by packing into the batch dimension, without altering the underlying computation. Cross-attention uses a single head at all layers, with per-layer head dimensions of $\{64, 48, 32, 16\}$.

\textbf{Self-attention.} After cross-attention, all segment states are stacked to form the segment-state matrix $\mathbf{E}^{(\ell)} \in \mathbb{R}^{B\times S\times d_\ell}$. Self-attention is applied across all $S$ segment states, enabling global information exchange:
\begin{equation}
\mathbf{E}^{(\ell)} \leftarrow \mathbf{E}^{(\ell)} + \mathrm{Attn}\bigl(\mathbf{E}^{(\ell)},\ \mathbf{E}^{(\ell)}\bigr),
\end{equation}
repeated $R_\ell \in \{8, 6, 4, 2\}$ times per layer for $\ell = 0,\ldots,3$. Self-attention uses multi-head attention with per-layer head counts of $\{8, 6, 4, 2\}$ and per-layer head dimensions of $\{64, 48, 32, 16\}$.

\textbf{Hierarchical segment-state compression.} Across the four layers, the segment-state dimensionality decreases progressively as $d_\ell \in \{128, 112, 96, 80\}$. At each layer transition, the segment representations are projected to the new dimensionality through a linear transformation when required. In the segmented-parallel configuration, the number of segment states remains fixed at $S=4$, with one state associated with each token segment. After the final layer, the final segment states $\mathbf{E}^{(4)} \in \mathbb{R}^{B\times S\times 80}$ are averaged across segments and passed through a linear classification head. Both attention operations use pre-layer normalization and residual connections, with attention and feedforward dropout of $0.10$ applied uniformly.

\begin{figure*}
\begin{center}
\includegraphics[scale=0.69]{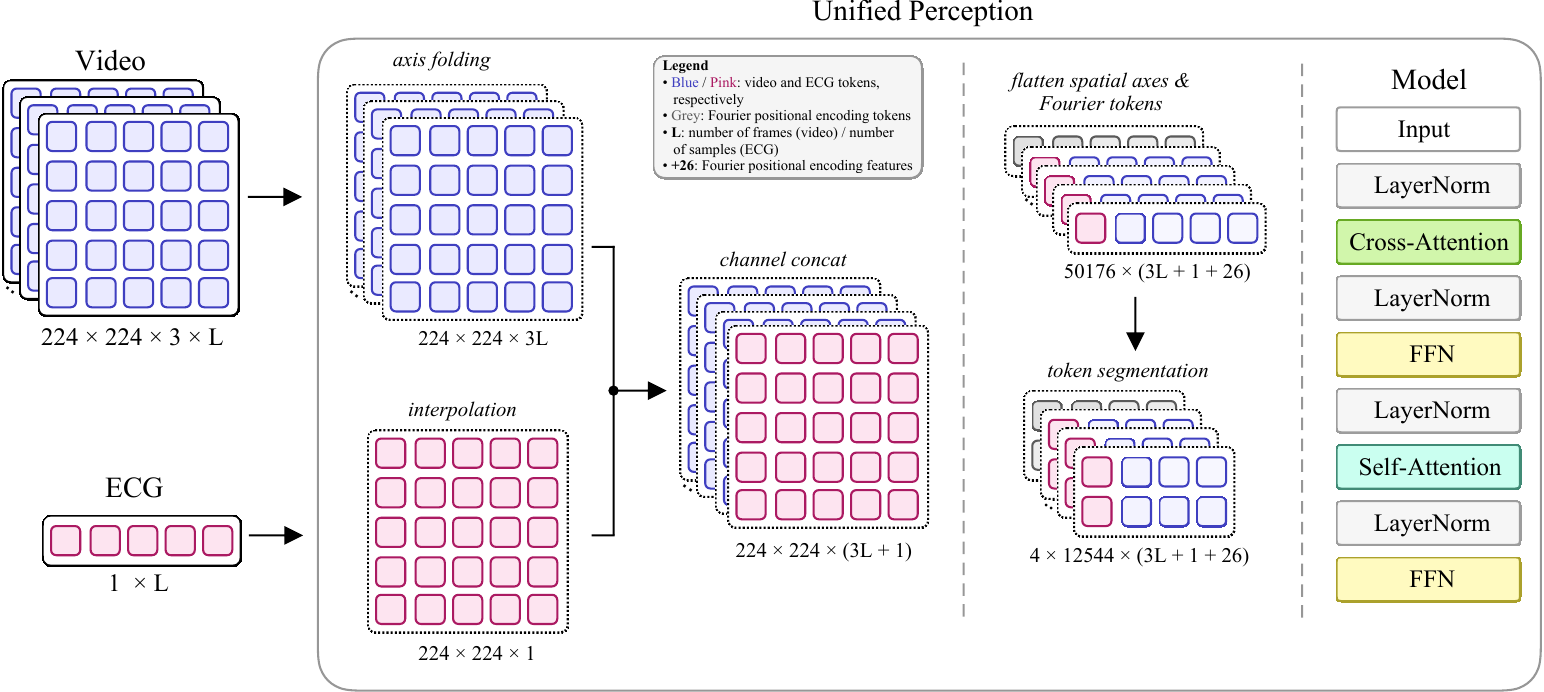}
\end{center}
\caption{Overview of the proposed unified perception framework. The video sequence is folded along the channel axis to form a $224\times224\times3L$ tensor, while the ECG waveform is interpolated to a $224\times224\times1$ grid. The two modalities are concatenated into a $224\times224\times(3L+1)$ input, flattened into $50176$ spatial tokens, augmented with $26$ Fourier positional features, and segmented into four token groups before being processed by the asymmetric attention model.}
\label{overview}
\end{figure*}

\section{Experimental Evaluation \& Results}
This section presents the experimental evaluation of MUPA\textsuperscript{2}S across unimodal and multimodal configurations. All experiments are evaluated under a binary classification setting. Validation performance is reported using macro-averaged accuracy, precision, and F1 score. Test performance is reported using macro-averaged accuracy.
All models are trained for $190$ epochs using AdamW with a learning rate of $10^{-4}$, weight decay of $0.05$, batch size $4$, and a cosine learning-rate schedule with $20$ warmup epochs. Cross-entropy loss with label smoothing of $0.15$ is used during training.

\subsection{Data Collection}
\label{ssec:data_collection}
In this study, a multimodal stress dataset comprising $58$ adults ($24$ men, $34$ women) aged $26.9\pm4.8$ years was collected. The experimental protocol was designed to investigate facial and physiological responses under stress through four induction categories: social exposure, emotional recall, mental workload, and stressful stimuli. Each participant completed $11$ tasks of $120$~sec each: social exposure (neutral reference, baseline description, and interview), emotional recall (neutral reference and recall of a stressful event), mental workload (reading reference, Stroop Colour-Word Test, and Paced Auditory Serial Addition Test (PASAT)), and stressful stimuli (relaxing video, adventure video, and psychological pressure). The interview, recall, Stroop, PASAT, adventure video, and psychological pressure tasks were stress-inducing, while the reference and description tasks were neutral. The relaxing video served as a physiological recovery baseline and was grouped with the neutral conditions for binary classification. The stress induction was validated both physiologically and subjectively, as heart rate increased significantly during the stress-inducing tasks ($p<0.05$), while the Self-Assessment Manikin ratings indicated elevated arousal and reduced valence in the stress phases compared with the neutral ones. It should be noted that the neutral tasks were administered at the beginning of each phase, and therefore the stress labels are partially aligned with task order. Moreover, the interview, recall, and PASAT tasks require the participants to speak, whereas the video stimuli do not. Consequently, part of the information learned from facial video may relate to speech-related facial activity, and the present design does not allow these factors to be fully separated. A detailed description of the protocol is provided in \cite{gkikas2026unwind}. The study received approval from the Research Ethics Committee (approval no.~155/12-09-2022). All participants provided informed consent. The dataset is available upon request for academic research purposes \footnote{\url{https://github.com/ggian/stress_dataset}}. Facial video and ECG were synchronously acquired during each task, with each sample comprising the paired Video--ECG recordings from the same experimental task. The facial video was recorded at $60$~fps and subsampled to $30$~fps with a resolution of $608\times800$ pixels, while ECG was recorded continuously on a single channel at $1$~kHz.

From the recorded $608\times800$ video frames, the face region was detected by \cite{bazarevsky_2019_blazeface} and resized to $224\times224$ before input to the model, providing a fixed spatial resolution for tokenization and matching the spatial domain used by the interpolated ECG representation in the unified fusion configuration.
No additional ECG filtering is applied; each ECG waveform is mean-centered, standardized, min--max normalized to $[0,1]$, and bilinearly interpolated to a $224\times224$ representation.
Binary classification is applied to distinguish neutral from stress conditions. Subjects are partitioned into training, validation, and testing sets at the subject level, ensuring no participant appears in more than one set. To avoid performance inflation due to subject-difficulty imbalance, a stratified split protocol is used. Leave-one-subject-out cross-validation is first conducted across all recorded modalities to estimate per-subject classification difficulty, providing a modality-independent ranking. Subjects are ranked by the combined z-score of these difficulty estimates and assigned to four quartile-based groups.
The final split comprises $38$ training, $8$ validation, and $12$ testing subjects, corresponding to $414$, $83$, and $132$ synchronized Video--ECG samples, respectively. The class distributions are $188/226$, $37/46$, and $60/72$ neutral/stress samples for training, validation, and testing, respectively, with each set containing subjects from all four groups in proportion.
Because the difficulty stratification is derived from preliminary LOSO estimates across the full cohort, we additionally evaluate the unified Video--ECG model using full LOSO across all $58$ participants, providing a complementary subject-independent evaluation that does not depend on the fixed train/validation/test partition.
 
\begin{table*}
\caption{Subject-level split by difficulty group. Subjects are ranked by combined z-score and assigned to four quartile-based groups (Q1\,=\, hardest, Q4\,=\, easiest).}
\label{tab:subject_split_stress}
 
\begin{center}
\begin{threeparttable}
\begin{tabular}{P{1.65cm} P{3.6cm} P{3.6cm} P{3.6cm} P{3.6cm}}
\toprule
 
\multirow[c]{3}{*}{Split} &
\multicolumn{4}{c}{Difficulty Group} \\
 
\cmidrule(lr){2-5}
 & Q1 -- Hard & Q2 -- Med-Hard & Q3 -- Med-Easy & Q4 -- Easy \\
 
\midrule
\midrule
Training (38) &
P017, P018, P022, P026, P034, P035, P042, P045, P050, P056 &
P001, P002, P003, P007, P012, P021, P033, P040, P048 &
P004, P014, P016, P032, P036, P046, P047, P052, P053, P054 &
P005, P010, P020, P028, P029, P037, P039, P041, P057 \\\hdashline
 
Validation (8) &
P038, P055 &
P009, P023 &
P006, P013 &
P019, P030 \\\hdashline
 
Testing (12) &
P008, P025, P044 &
P011, P024, P043 &
P015, P031, P058 &
P027, P051, P059 \\
 
\bottomrule
\end{tabular}
\begin{tablenotes}[para,flushleft]
\scriptsize
\item Q1: $z < -0.46$;\quad Q2: $-0.46 \leq z < -0.05$;\quad Q3: $-0.05 \leq z < +0.40$;\quad Q4: $z \geq +0.40$.
\end{tablenotes}
\end{threeparttable}
\end{center}
\end{table*}

\subsection{Unimodal Video}
\label{sec:video}

For video, temporal stride $\tau$ denotes uniform frame subsampling, where every $\tau$-th frame is retained; thus, an original sequence of $T$ frames produces $L=\lfloor T/\tau \rfloor$ retained frames. Table~\ref{table:videos} reports performance and computational cost across five temporal stride configurations.
Test accuracy ranges from $60.56\%$ at stride~$5$ to $69.44\%$ at stride~$15$, with stride~$15$ achieving the highest test performance. The validation peak occurs at stride~$20$ ($70.42\%$), showing that validation and test performance do not peak at the same temporal resolution. GFLOPs decrease substantially with stride, from $70.46$ at stride~$5$ to $12.48$ at stride~$30$, reflecting the reduction in the folded video-channel dimension as fewer frames are retained, while the spatial token count remains fixed at $224\times224=50176$. Parameters also decrease slightly across configurations, from $6.39$M to $5.82$M, as the number of input channels scales with the number of frames and directly affects the size of the cross-attention projection matrices. The results show that test accuracy varies across temporal strides, with the best performance obtained at stride~$15$, indicating that retaining more frames does not necessarily improve performance.
The stride analysis nevertheless shows that intermediate temporal sampling provides a favorable trade-off between predictive performance and computational cost. Notably, at stride~$5$ the model processes $720$ frames as a single complete input tensor without any external windowing. This highlights that the architecture can natively handle full-length video inputs via its internal token segmentation mechanism.

\subsection{Unimodal ECG}
\label{sec:ecg}

Table~\ref{table:ecg} reports performance and computational cost across seven stride configurations for the ECG modality. 
Test accuracy ranges from $51.81\%$ to $53.33\%$ across all configurations, remaining close to the chance level throughout. Stride~$15$ achieves the marginally highest test accuracy ($53.33\%$), though the differences across strides are negligible. Unlike the video modality, both GFLOPs and parameter count are identical across all stride configurations at the reported precision ($1.12$ and $5.84$M, respectively), as the 1D ECG token sequences are short and the computational differences across strides are negligible at two decimal places. Under the proposed unified representation, ECG-only performance remains close to chance, indicating limited discriminative performance for the ECG modality within the present modeling configuration.

\begin{table*}
\caption{Performance and computational cost using the video modality.}
\label{table:videos}

\begin{center}
\begin{threeparttable}
\begin{tabular}{P{1.5cm} P{0.7cm} P{1.4cm} P{1.1cm}  P{0.9cm} P{0.9cm} P{0.4cm} P{1.0cm}}
\toprule

\multicolumn{2}{c}{Input} &
\multicolumn{2}{c}{Computational Cost} &
\multicolumn{3}{c}{Validation} &
\multicolumn{1}{c}{Testing} \\

\cmidrule(lr){1-2}\cmidrule(lr){3-4}\cmidrule(lr){5-7}\cmidrule(lr){8-8}
Modality &Stride & Params (M) & GFLOPs & Accuracy & Precision & F1 &Accuracy  \\

\midrule
\midrule
Video & 5  &6.39 &70.46 &68.80 &69.39 &67.39 &60.56 \\\hdashline
Video & 10 &6.05 &35.67 &66.39 &66.20 &66.14 &63.75 \\\hdashline
Video & 15 &5.93 &24.07 &67.77 &68.33 &67.89 &69.44 \\\hdashline
Video & 20 &5.87 &18.27 &70.42 &71.86 &68.45 &65.97 \\\hdashline
Video & 30 &5.82 &12.48 &68.60 &69.91 &68.72 &64.86
\\

\bottomrule
\end{tabular}
\begin{tablenotes}[para,flushleft]
\scriptsize

\end{tablenotes}
\end{threeparttable}
\end{center}
\end{table*}

\begin{table*}
\caption{Performance and computational cost using the ECG modality.}
\label{table:ecg}

\begin{center}
\begin{threeparttable}
\begin{tabular}{P{1.5cm} P{0.7cm} P{1.4cm} P{1.1cm}  P{0.9cm} P{0.9cm} P{0.4cm} P{1.0cm}}
\toprule

\multicolumn{2}{c}{Input} &
\multicolumn{2}{c}{Computational Cost} &
\multicolumn{3}{c}{Validation} &
\multicolumn{1}{c}{Testing} \\

\cmidrule(lr){1-2}\cmidrule(lr){3-4}\cmidrule(lr){5-7}\cmidrule(lr){8-8}
Modality &Stride & Params (M) & GFLOPs & Accuracy & Precision & F1 &Accuracy  \\

\midrule
\midrule
ECG & 1  &5.84 &1.12 &50.00 &27.71 &35.66 &51.81 \\\hdashline
ECG & 2  &5.84 &1.12 &54.32 &68.85 &45.81 &52.50 \\\hdashline
ECG & 5  &5.84 &1.12 &55.93 &66.83 &49.70 &52.50 \\\hdashline
ECG & 10 &5.84 &1.12 &55.67 &70.89 &48.21 &52.64 \\\hdashline
ECG & 15 &5.84 &1.12 &53.23 &61.90 &45.07 &53.33 \\\hdashline
ECG & 20 &5.84 &1.12 &55.67 &70.89 &48.21 &52.64 \\\hdashline
ECG & 30 &5.84 &1.12 &54.58 &64.66 &47.43 &52.78
\\

\bottomrule
\end{tabular}
\begin{tablenotes}[para,flushleft]
\scriptsize
\item Reported GFLOPs are identical across all stride configurations at the displayed precision. The 1D ECG token sequences are short relative to the video representation, and the GFLOP differences across strides are negligible at two decimal places.
\end{tablenotes}
\end{threeparttable}
\end{center}
\end{table*}

\subsection{Multimodal Video \& ECG}
Two fusion strategies are evaluated and compared against the unimodal baselines reported in Sections~\ref{sec:video} 
and~\ref{sec:ecg}. The first is a feature-level approach, in which video and ECG are processed independently by two instances of the same model architecture. The final latent features from the two branches are pooled and concatenated, and the resulting joint representation is passed through a classification fusion head. The second is the unified approach described in Section~\ref{sec:methodology}, where both modalities are combined at the input level through channel 
concatenation and processed by a single shared model. These two strategies are evaluated to compare the accuracy and 
efficiency of using separate model instances against a single unified model. For all multimodal configurations, the ECG stride is fixed at $1$. Since the ECG waveform is projected onto a fixed $224\times224$ spatial grid through bilinear interpolation before tokenization, the ECG input size remains fixed regardless of the original signal length. The stride notation $x$-$y$ in Table~\ref{table:multimodal} denotes the video stride $x$ and the ECG stride $y$.

\begin{figure}
\begin{center}
\includegraphics[scale=0.55]{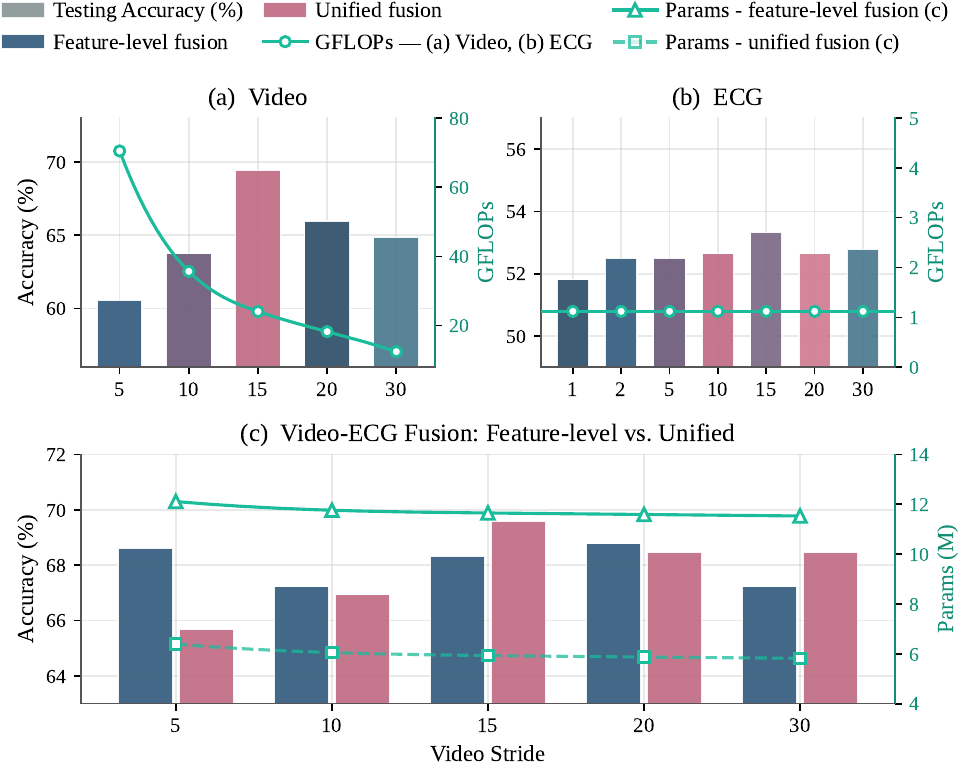}
\end{center}
\caption{Comparison of test accuracy and computational cost across unimodal and multimodal configurations. Panels (a) and (b) report video-only and ECG-only performance across stride values, together with the corresponding GFLOPs. Panel (c) compares feature-level and unified Video-ECG fusion across video strides, with ECG stride fixed to $1$. Unified fusion achieves an overall test accuracy similar to feature-level fusion while using approximately half as many parameters.}
\label{comparison}
\end{figure}

Table~\ref{table:multimodal} and Fig.~\ref{comparison}(c) report the results for both fusion strategies across five 
video stride configurations. Feature-level fusion achieves test accuracies between $67.22\%$ and $68.78\%$, with the highest test performance obtained at stride~$20$ ($68.78\%$). Compared with the video-only results, feature-level fusion improves performance when we use low strides, with the largest gain at stride~$5$ where test accuracy increases from $60.56\%$ to $68.61\%$, a gain of approximately $8$ percentage points. However, feature-level fusion does not improve over the best video-only configuration ($69.44\%$ at stride~$15$), indicating that combining independently trained representations provides robustness at configurations where video-only performance is lower. The number of parameters for feature-level fusion ranges from $11.53$M to $12.11$M, reflecting the cost of maintaining two model instances.

The unified fusion strategy achieves test accuracies between $65.69\%$ and $69.58\%$, with the highest test performance obtained at stride~$15$ ($69.58\%$). The unified model uses approximately half as many parameters as feature-level fusion across all strides ($5.82$M to $6.39$M), since it processes both modalities within a single shared backbone. GFLOPs for the unified approach remain nearly identical to the corresponding video-only configurations, since the ECG channel adds only one additional input channel after interpolation and contributes negligible computational overhead. 
Feature-level fusion adds approximately $0.91$~GFLOPs over its video-only counterpart across all stride configurations.
Overall, unified fusion achieves the highest test accuracy with the lowest parameter count among multimodal configurations, while feature-level fusion yields more consistent performance across stride configurations at approximately twice the parameter cost.

\begin{table*}
\caption{Performance and computational cost using the video and ECG modalities.}
\label{table:multimodal}

\begin{center}
\begin{threeparttable}
\begin{tabular}{P{1.5cm} P{0.7cm} P{2.0cm} P{1.4cm} P{1.1cm}  P{0.9cm} P{0.9cm} P{0.4cm} P{1.3cm}}
\toprule

\multicolumn{3}{c}{Input} &
\multicolumn{2}{c}{Computational Cost} &
\multicolumn{3}{c}{Validation} &
\multicolumn{1}{c}{Testing} \\

\cmidrule(lr){1-3}\cmidrule(lr){4-5}\cmidrule(lr){6-8}\cmidrule(lr){9-9}
Modality &Stride &Fusion & Params (M) & GFLOPs & Accuracy & Precision & F1 &Accuracy  \\

\midrule
\midrule
Video-ECG & 5-1  &feature-level &12.11 &71.37 &69.36 &69.27 &68.67 &68.61\\\hdashline
Video-ECG & 10-1 &feature-level &11.76 &36.58 &70.48 &70.74 &70.57 &67.22\\\hdashline
Video-ECG & 15-1 &feature-level &11.65 &24.99 &71.27 &71.02 &70.98 &68.33 \\\hdashline
Video-ECG & 20-1 &feature-level &11.59 &19.19 &67.74 &67.54 &67.39 &68.78 \\\hdashline
Video-ECG & 30-1 &feature-level &11.53 &13.39 &68.57 &68.41 &68.45 &67.22 \\\midrule

Video-ECG & 5-1  &unified &6.39 &70.49 &66.13 &66.00 &66.02 &65.69 \\\hdashline
Video-ECG & 10-1 &unified &6.05 &35.70 &66.92 &66.84 &66.26 &66.94 \\\hdashline
Video-ECG & 15-1 &unified &5.93 &24.11 &65.83 &65.83 &65.03 &69.58 \\\hdashline
Video-ECG & 20-1 &unified &5.87 &18.31 &69.33 &71.01 &67.16 &68.47 \\\hdashline
Video-ECG & 30-1 &unified &5.82 &12.51 &67.98 &69.25 &66.02 &68.47
\\

\bottomrule
\end{tabular}
\begin{tablenotes}[para,flushleft]
\scriptsize
\item The stride notation $x$-$y$ denotes the video stride $x$ and the ECG stride $y$. ECG stride is fixed at $1$ across all configurations, as the waveform is projected onto a fixed $224\times224$ spatial grid through bilinear interpolation regardless of its original length.

\end{tablenotes}
\end{threeparttable}
\end{center}
\end{table*}

\subsection{Overall Analysis \& Discussion}

Across unimodal configurations, facial video reaches $69.44\%$ test accuracy at stride~$15$, while ECG ranges from $51.81\%$ to $53.33\%$; these results characterize performance within the proposed framework rather than intrinsic modality informativeness. Unified Video--ECG fusion achieves $69.58\%$ test accuracy at stride~$15$, compared with $68.78\%$ for feature-level fusion, while requiring approximately half the parameters.
To complement the fixed hold-out protocol, unified Video--ECG fusion (video stride~$15$, ECG stride~$1$) was evaluated using LOSO across all $58$ participants, achieving $80.30\pm11.63\%$ accuracy, $82.59\pm12.06\%$ precision, and $78.92\pm13.24\%$ F1 score, independently of the difficulty-based partition. Table \ref{table:comparison} compares MUPA\textsuperscript{2}S with three approaches evaluated on the same $58$-subject dataset. Under stratified hold-out, EDA-based representation fusion achieves $70.97\%$ test accuracy, followed by MUPA\textsuperscript{2}S ($69.58\%$) and \cite{gkikas2026fuse} ($69.44\%$); under LOSO, MUPA\textsuperscript{2}S reaches $80.30\pm11.63\%$, close to the $80.04\pm10.86\%$ reported in \cite{gkikas2026unwind}.
Although the best unified and video-only configurations differ by only $0.14$ percentage points, unified fusion improves over video alone by $5.13$, $3.19$, $0.14$, $2.50$, and $3.61$ points at matched strides of $5$, $10$, $15$, $20$, and $30$, respectively. While these results do not establish a statistically significant advantage over ECG, they suggest that its integration can preserve or improve performance across the evaluated configurations while providing complementary sensing redundancy for health-related assessment in clinical environments.

\begin{table}
\caption{Comparison with previous approaches evaluated on the same $58$-subject stress dataset.}
\label{table:comparison}

\begin{center}
\begin{threeparttable}
\scriptsize
\setlength{\tabcolsep}{3pt}

\begingroup

\begin{tabular}{P{1.5cm} P{1.5cm} P{1.4cm} P{2.0cm}}
\toprule

Study & Modality & Protocol & Accuracy \\

\midrule
\midrule

\cite{gkikas2026edastress}
& EDA
& Hold-out
& 70.97 \\\hdashline

\cite{gkikas2026fuse}
& Video
& Hold-out
& 69.44 \\\hdashline


MUPA\textsuperscript{2}S
& Video--ECG
& Hold-out
& 69.58 \\\midrule

\cite{gkikas2026unwind}
& Video
& LOSO
& $80.04 \pm 10.86$ \\\hdashline

MUPA\textsuperscript{2}S
& Video--ECG
& LOSO
& $80.30 \pm 11.63$ \\

\bottomrule
\end{tabular}

\endgroup

\begin{tablenotes}[para,flushleft]
\scriptsize
\item Hold-out denotes the same stratified subject-level evaluation protocol. LOSO results are reported as mean accuracy $\pm$ standard deviation.
\end{tablenotes}

\end{threeparttable}
\end{center}
\end{table}

\section{Conclusion}

This paper presented MUPA\textsuperscript{2}S, a multimodal unified perception framework for automatic stress assessment combining facial video and ECG signals. The framework maps both modalities into a shared representation and processes them through asymmetric attention, enabling heterogeneous inputs to be handled within a single architecture without modality-specific components.
Experiments on a newly collected $58$-subject stress dataset show that facial video reaches $69.44\%$ test accuracy, while unified Video--ECG fusion achieves the highest observed test accuracy of $69.58\%$ with approximately half the parameters of feature-level fusion. Complementary LOSO evaluation achieves $80.30\pm11.63\%$ accuracy, $82.59\pm12.06\%$ precision, and $78.92\pm13.24\%$ F1 score. 
We emphasize that the present study is based on a single controlled dataset, and in real-world settings this could be limited.
We suggest that future work should evaluate additional datasets, either stress-based or other internal human states \cite{gkikas_reface_acii_2026} that can influence stress sensation or manifestation, while exploring richer physiological representations \cite{kyprakis_gkikas_acii_2026} and complementary modalities \cite{cruz_gkikas_emotion_acii_2026} to improve generalization and sensing reliability in health-related applications  \cite{gkikas_cruz_pain_eeite_2026}, such as diagnostic systems or companion robots \cite{montiel_cruz_gkikas_eeite_2026} in healthcare settings.

\section*{Acknowledgments}
The authors used large language model (LLM)-based tools for language editing and improvement. All scientific content, results, and conclusions are solely the work of the authors.


\bibliographystyle{IEEEtran}
\bibliography{library}

\end{document}